\documentclass{article}

\usepackage{arxiv}

\usepackage[utf8]{inputenc} % allow utf-8 input
\usepackage{graphicx}
\usepackage[T1]{fontenc}    % use 8-bit T1 fonts
\usepackage{hyperref}       % hyperlinks
\usepackage{url}            % simple URL typesetting
\usepackage{booktabs}       % professional-quality tables
\usepackage{amsfonts}       % blackboard math symbols
\usepackage{nicefrac}       % compact symbols for 1/2, etc.
\usepackage{microtype}      % microtypography
\usepackage{lipsum}		% Can be removed after putting your text content
\usepackage{array}
\usepackage{multirow}
\usepackage{geometry}
\usepackage{fancyhdr}
\title{Safety experiments for small robots investigating the potential of soft materials in mitigating the harm to the head due to impacts}

\author{
  Ahmad Yaser Alhaddad\\
  Department of Mechanical and Industrial Engineering\\
  Qatar University, Doha 2713, Qatar\\
  Department of Electronics, Information and Bioengineering \\
  Politecnico di Milano, Milano 20133, Italy\\
  \texttt{} \\
   \And
 John-John Cabibihan\thanks{PLEASE CITE THIS ARTICLE IN PRESS AS: Ahmad Yaser Alhaddad, John-John Cabibihan, Ahmad Hayek, Andrea Bonarini, "Safety experiments for small robots investigating the potential of soft materials in mitigating the harm to the head due to impacts," SN applied Sciences, Springer Nature, 2019, 1: 476, doi 10.1007/s42452-019-0467-7} \\
  Department of Mechanical and Industrial Engineering\\
  Qatar University, Doha, Qatar\\
  \texttt{john.cabibihan@qu.edu.qa} \\
   \And
  Ahmad Hayek \\
  Department of Mechanical and Industrial Engineering\\
  Qatar University, Doha, Qatar\\
  \And
  Andrea Bonarini \\
  Department of Electronics, Information and Bioengineering \\
  Politecnico di Milano, Milano 20133, Italy\\
}

\begin{document}
\maketitle

\begin{abstract}
There is a growing interest in social robots to be considered in the therapy of children with autism due to their effectiveness in improving the outcomes. However, children on the spectrum exhibit challenging behaviors that need to be considered when designing robots for them. A child could involuntarily throw a small social robot during meltdown and that could hit another person's head and cause harm (e.g. concussion). In this paper, the application of soft materials is investigated for its potential in attenuating head's linear acceleration upon impact. The thickness and storage modulus of three different soft materials were considered as the control factors while the noise factor was the impact velocity. The design of experiments was based on Taguchi method. A total of 27 experiments were conducted on a developed dummy head setup that reports the linear acceleration of the head. ANOVA tests were performed to analyze the data. The findings showed that the control factors are not statistically significant in attenuating the response. The optimal values of the control factors were identified using the signal-to-noise (S/N) ratio optimization technique. Confirmation runs at the optimal parameters (i.e. thickness of 3 mm and 5 mm) showed a better response as compared to other conditions. Designers of social robots should consider the application of soft materials to their designs as it help in reducing the potential harm to the head.
\end{abstract}

% keywords can be removed

\section{Introduction}
The interest in robots is increasing globally as estimated by the International Federation of Robotics (IFR) \cite{International2016executive}. The application of robots is extending to new areas, such as that in healthcare. Most notably is the application of social robots in therapy sessions with children with autism, which has been reported to improve the overall outcomes \cite{cabibihan2013robots}. However, such children exhibit a multitude of challenging behaviors that could raise some safety concerns when a robot is present in their vicinity \cite{Alhaddad2018}. The occurrence rates of challenging behaviors are high (e.g. 49\% up to 69\% \cite{baghdadli2003risk}\cite{bodfish2000varieties}\cite{kanne2011aggression}), and that have many consequences on the services and treatments provided to them \cite{hutchins2014using}.       

\begin{figure}[t]
	\centering
	\includegraphics[width=1\textwidth]{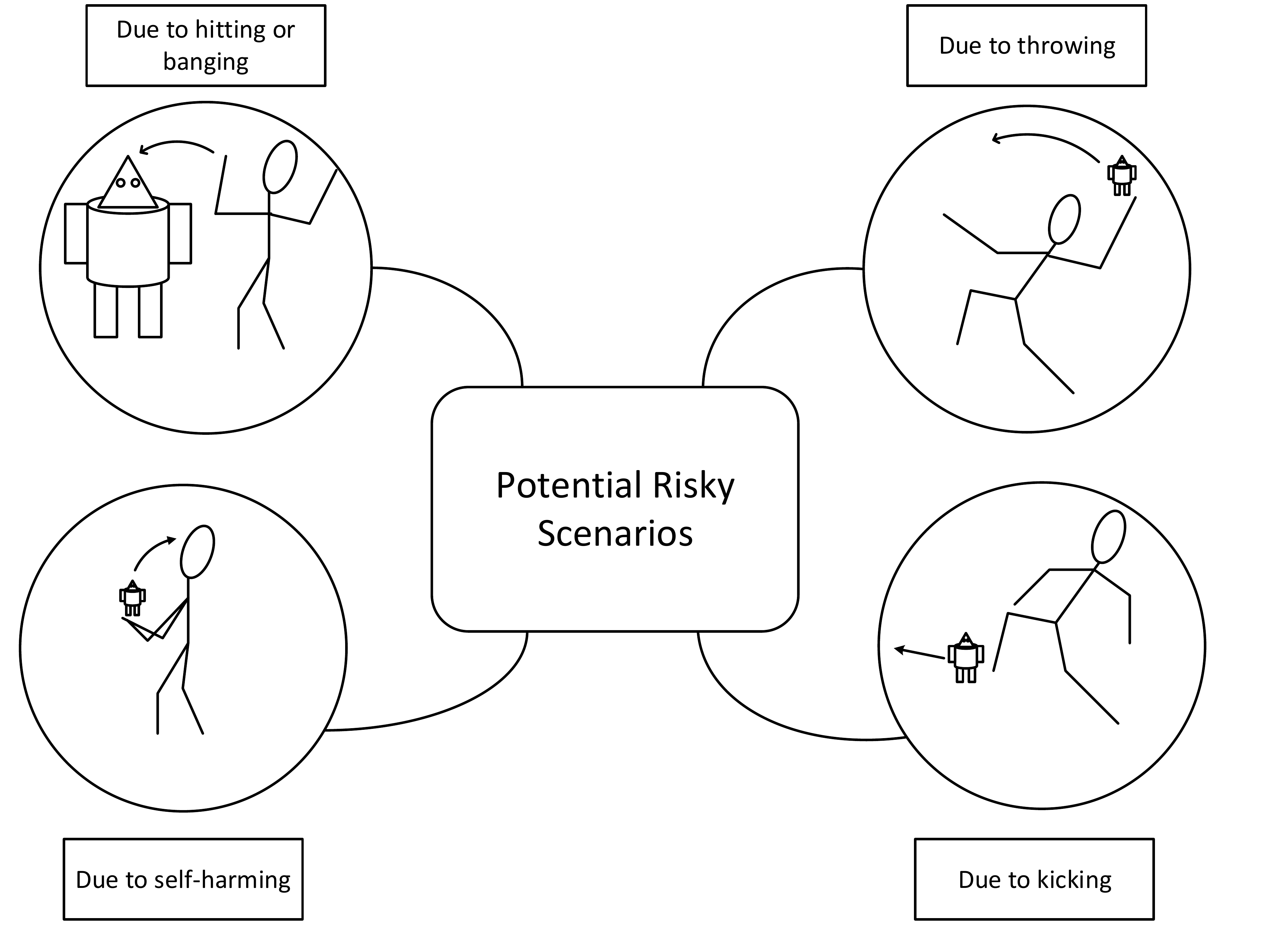}
	\caption{The identified possible risky scenarios that might occur between a child and social robot due to some of the challenging behaviors \cite{Alhaddad2018}.}
	\label{Drawing_scenarios}
\end{figure}

The ability to convey emotions, exhibiting different personalities, employing communication cues, and forming social relationships are some of the traits that make social robots different from typical toys \cite{so2016using}\cite{cabibihan2009pointing}\cite{wykowska2015humans}. Introducing a new stimuli (e.g. a social robot) that is meant to elicit behaviors could provoke some unwanted behaviors among children with autism. For example, they might show some aggression toward the robots \cite{alhaddad2018robotic}\cite{boccanfuso2016emotional}\cite{Cabibihan2018}. Kicking, throwing, hitting, and banging are some of the challenging behaviors that could potentially cause harm during the interactions with social robots (Fig. \ref{Drawing_scenarios}). For example, the throwing of a small social robot to the head could cause superficial injuries, subconcussion or even concussion in extreme cases \cite{Alhaddad2018}. The occurrence of such behaviors implies the need for safer robotic designs \cite{alhaddad2018robotic}\cite{cabibihan2018smart}.

The establishment of safety standards in different fields of robotics is making a notably advances. However, the progress in establishing safety standards in relation to social robots and robotic toys is still lacking \cite{herrmann2010towards}\cite{diep2014social}. Some of the existing safety standards in toys can be readily imported to cover some fundamental design and safety aspects. For example, the ISO 8124 standard \cite{safety2014toys}. Safety aspects of the mechanical and physical properties of toys are covered in part one of this standard while part two and three covers flammability and migration of certain elements, respectively. More rigorous design considerations are needed that consider the unwanted behaviors exhibited by children with autism. For example, considering a scenario where the robotic toy is thrown to the head.       

In this paper, we use Taguchi design of experiment (DOE) method to investigate  a way to reduce the harm to the head by studying the influence of two control factors (i.e. soft material storage modulus and its thickness) and one noise factor (i.e. throwing velocity) of a small form factor toy on the resultant head's acceleration. Furthermore, the optimal levels of the investigated control factors that help in reducing the response are identified. This paper is organized as follows. Section 2 describes the Taguchi method while Section 3 provides the materials and methods. Section 4 presents the results of the study. Section 5 provides discussion while Section 6 concludes the work.

\section{Taguchi Method}

\subsection{Background}
The understanding of a process is usually done through conducting a series of experiments that provide information about the processes, such as the most influential parameters, the optimal settings, and the overall performance \cite{montgomery2017design}. To maximize the information obtained while minimizing the resources needed, design of experiment (DOE) approach is usually used to study processes. Taguchi's method is one of the DOE techniques that was developed in 1979 to improve the quality of goods \cite{taguchi1979introduction}. This technique has
been considered in the optimization of different applications, such as strain measurements, wear studies, and turning parameters \cite{Moayyedian2018}\cite{Hebbale2019}\cite{Tamizharasan2019}.  
% This technique has been considered in many applications.  

There are two types of process factors that are considered in Taguchi method. The first factor is controllable at the product design level and referred to as the control factor. The other factor is known as the noise factor that is uncontrollable during production and at product design level. However, noise factor can be simulated experimentally. Taguchi method aims to make products more robust by optimizing the control factors to reduce the sensitivity to noise factors. Furthermore, Taguchi method provides the quality loss function to measure losses in relation to product’s variation from the target performance. In Taguchi DOE, different Orthogonal Arrays (OAs) can be considered depending on the number of factors being investigated \cite{mori2011taguchi}.

\begin{table}
	\centering
	\caption{The standard $L_{9}(3^{2})$ orthogonal array (OA) that was used in this study.}{
		
		\begin{tabular}{>{\centering\arraybackslash}p{2cm}>{\centering\arraybackslash}p{2.5cm}>{\centering\arraybackslash}p{2cm}} \toprule Run & \multicolumn{2}{c}{Control factors} \\
			& A & B \\\midrule
			1 & 1 & 1 \\
			2 & 1 & 2 \\
			3 & 1 & 3 \\
			4 & 2 & 1 \\
			5 & 2 & 2 \\
			6 & 2 & 3 \\
			7 & 3 & 1 \\
			8 & 3 & 2 \\
			9 & 3 & 3 \\\bottomrule
	\end{tabular}}
	\label{OA_9}
\end{table}

\subsection{Experimental design}
Taguchi design method has been considered in this study because it considers the noise variables and because it greatly reduces the number of experiments. Furthermore, the method provides a robust parameter design by finding the optimal values of the control variables to reduce the investigated problem sensitivity to noise variables \cite{cavazzuti2012optimization}. The Orthogonal Arrays (OAs) of this method provides more information to understand the relationship between the control variables and the noise variables, which is essential for a robust design. Additionally, Taguchi method provides performance measures that are needed in this study to optimize the design, such as the signal-to-noise (S/N) ratio.

In this study, two control factors are investigated for their potential in reducing the linear acceleration of the head. Experiments are conducted based on an $L_{9}(3^{2})$ Taguchi OA (Table \ref{OA_9}). A total of 27 (i.e. $9 \times 3$) experiments have to be conducted that consider the three levels of the control and noise factors. The considered control factors can be adjusted at the product design level while the noise factor is dependent on the real life scenario (e.g. throwing). Finally, the selected factors are independent while the measured output (i.e. head's acceleration) is dependent.

\begin{table*}[]
	\centering
	\caption{The factors and their levels that have been considered in this study. The values of the noise factor (i.e. impact velocity) were divided into three different ranges. The object was dropped at three different heights corresponding to three different ranges of impact velocities. Based on the analysis of the videos, the intervals have been established. Such intervals have been used as levels to investigate the contribution of the noise factor on the response. }{
		\begin{tabular}{p{2cm}lllll} \toprule
			Type & Parameter & Code & Level 1 & Level 2 & Level 3 \\\midrule
			\multirow{2}{*}{Control factor} & \multirow{2}{*}{Thickness (mm)} & \multirow{2}{*}{A} & \multirow{2}{*}{1} & \multirow{2}{*}{3} & \multirow{2}{*}{5} \\
			& & & & & \\
			\multirow{1}{*}{Control factor}& Storage modulus (MPa) & B & 0.2 & 0.3 & 1.7 \\
			\multirow{2}{*}{Noise factor} & \multirow{2}{*}{Impact velocities (m/s)} & \multirow{2}{*}{X} & \multirow{2}{*}{Low} & \multirow{2}{*}{Medium} & \multirow{2}{*}{High } \\
			& \multirow{2}{*}{}& & \multirow{2}{*}{(1.0-1.2)}& \multirow{2}{*}{(1.6-2.0)} & \multirow{2}{*}{(2.6-3.0)}\\
			\multirow{3}{*}{Response} & \multirow{3}{*}{Peak linear head} \\ & \multirow{3}{*}{} & & & & \\
			& acceleration (g)&\\
			\\\bottomrule
	\end{tabular}}
	\label{variables}
\end{table*}

The levels of the two control factors (i.e. material thickness and storage modulus) and noise factor (i.e. impact velocity) have been defined (Table \ref{variables}). To achieve consistency, the mass and the shape of the impactor were kept the same throughout the experiments. The mass of the impactor was kept at 0.4 kg, which is within the expected range of the targeted applications (i.e. small robotic toys). The shape of the impactor was cylindrical without any features on the surface. Finally, the impact velocities used were limited to low velocities to achieve more consistency in terms of the noise levels (i.e. less than 3 m/s) \cite{moss2014towards}.

\begin{figure}[]
	\centering
	\includegraphics[width=0.75\textwidth]{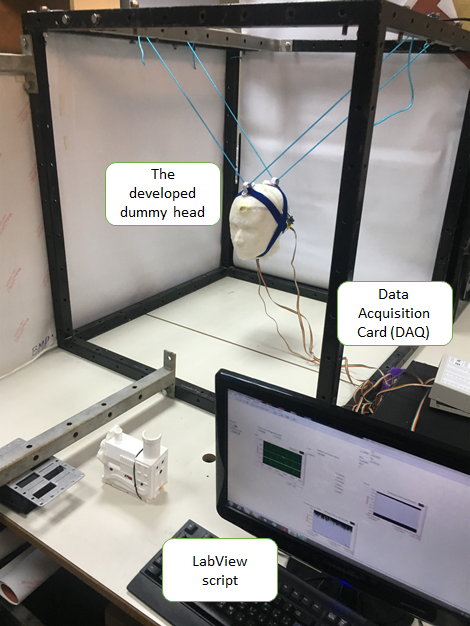}
	\caption{The experimental setup used in this study \cite{Alhaddad2018}.}
	\label{Expt_Setup}
\end{figure}

\section{Materials and Methods}

\subsection {Experimental Setup}

The experimental setup was based on a 3D printed dummy head situated with nylon coated wire ropes in a dedicated frame (Fig. \ref{Expt_Setup}). The head was embedded with an accelerometer that measures the linear acceleration of the head. The accelerometer sensor was interfaced to the computer through a data acquisition card. The experimental setup has been validated in giving comparable results to those of previous studies under similar testing conditions. More detailed description of the experimental setup can be viewed in our earlier studies \cite{Alhaddad2018}\cite{Alhaddad2019}\cite{DVN/AVC8GG_2018}.    

\subsection {Impactor}
\begin{figure}[]
	\centering
	\includegraphics[width=0.65\textwidth]{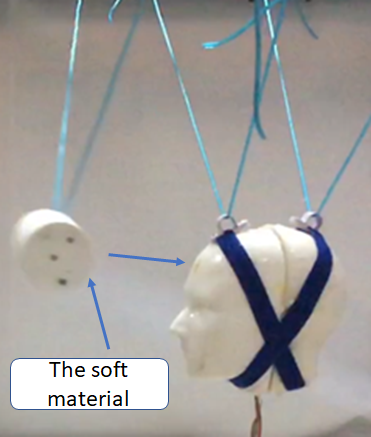}
	\caption{Sample of the experiments conducted.}
	\label{Sample_exp}
\end{figure}

A 3D printed cylindrical object was used as an impactor in this study. The dimensions of the impactor were ($10 \times 10$ $cm^{2}$, height and diameter). A 3D printer (Replicator 5$^{th}$ Generation, MakerBot Industries, USA) was used to build the object. Clay was used to fill the impactor to reach 0.4 kg. The soft materials (Ecoflex OO-30 \& Dragon skin FX‑ Pro, Smooth-On, USA) were prepared according to manufacturer's instructions. The soft materials were prepared in molds of different thicknesses and then rectangular ($5 \times 8$ $cm^{2}$) samples of each were attached to the impactor covering the area of impacts (Fig. \ref{Sample_exp}). 

\subsection {Procedures}
\subsubsection{Dynamic Mechanical Analysis}
\begin{figure}[]
	\centering
	\includegraphics[width=0.65\textwidth]{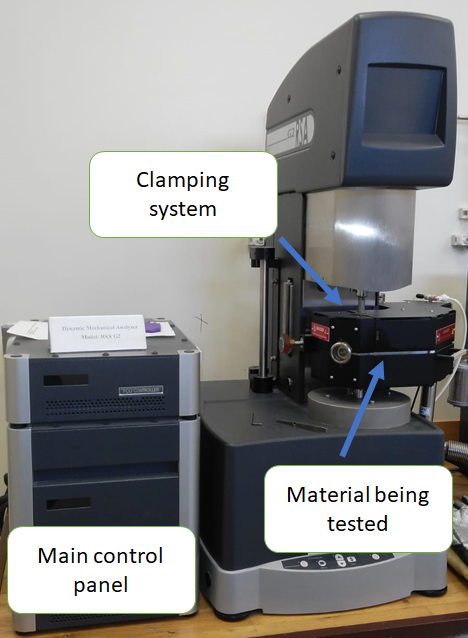}
	\caption{The Dynamic Mechanical Analyzer device that was used to analyze the soft materials.}
	\label{DMA_device}
\end{figure}

\begin{figure}[]
	\centering
	\includegraphics[width=0.90\textwidth]{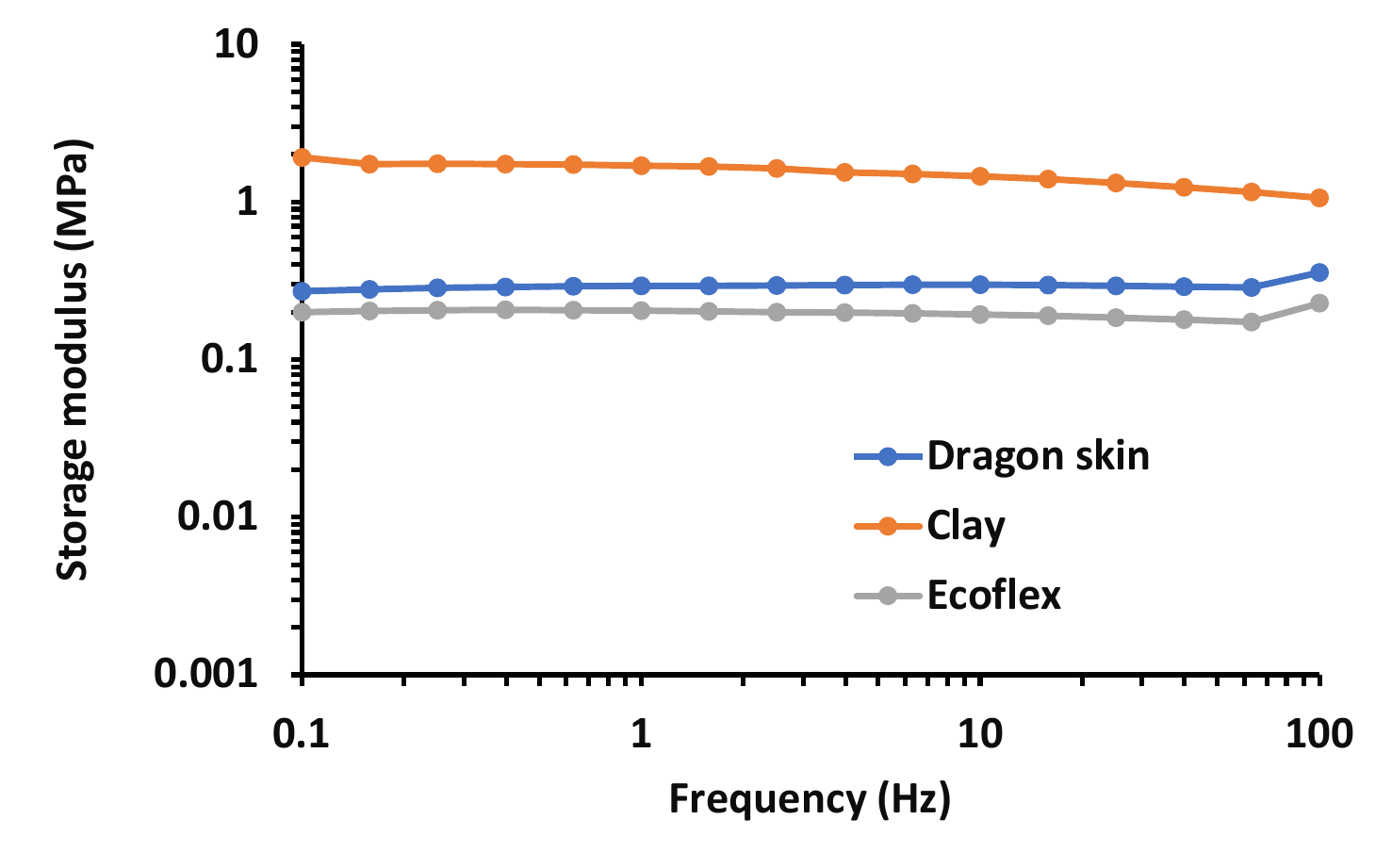}
	\caption{The storage modulus results for the dynamic mechanical analysis (DMA) tests that were performed on the three materials.}
	\label{DMA_results}
\end{figure}

The soft materials properties were studied using a dynamic mechanical analyzer (RSA-G2, TA instruments, USA ; Fig. \ref{DMA_device}). The dynamic mechanical analysis (DMA) is a common test to measure the properties (i.e. elastic and viscous) of a material. The properties were studied by applying a stress (e.g. sinusoidal) and measuring the resultant strain and the phase difference between the input and output. A frequency sweep tests were conducted to study the storage modulus. In these tests, the frequency was varied from 0.1 Hz to 100 Hz while the strain and temperature kept constant. The storage modulus readings for each material were generated (Fig. \ref{DMA_results}). The values of storage modulus at 1 Hz for each material were considered in the analysis. This frequency is believed to be at which the high rate of challenging behaviors might occur \cite{plotz2012automatic}.  

\subsubsection{Impact Experiments}
A total of 27 main experiments (i.e. 9 experiments at each noise level) were conducted  (See \href{https://youtu.be/rFM08uFQCaM}{supplementary material} and data \cite{Alhaddad2019data}\cite{DVN/TXOPUH_2019}).The experiments covered different combinations of the control factors according to the $L_{9}(3^{2})$ orthogonal array (OA) (Table \ref{OA_9}). To achieve consistent impact velocities (i.e. noise levels), the impactor was tied to the frame allowing it to swing freely (Fig. \ref{Sample_exp}). By altering the drop height, three different impact velocities levels were generated in the range of 1 - 3 m/s (Table \ref{variables}). As for the confirmation runs, a total of 24 runs at each noise level were conducted after identifying the optimal control factors levels from the main experiments.

A video camera (FDR-X1000V, Sony, Japan) was used to record all the experiments in slow-motion mode (240 fps, 720 pixels). An open-source video analysis software (Tracker version 4.10.0, Douglas Brown, Open Source Physics) was used to analyze the videos for impact velocities. A script based on LabView (2014, National Instrument, USA) was used to acquire and then save the raw data. A Matlab (Version 2015, MathWorks, Massachusetts, USA) script was used to post-process the data for the peak acceleration values.

\subsection{Peak Linear Head Acceleration}
The peak linear acceleration of the head during impacts has been used as one of the biomechanical measures to investigate concussive events \cite{greenwald2008head}\cite{rowson2013brain}. In a study on young football players (age 7 - 8 years) \cite{young2014head}, the reported peak linear accelerations during impacts could fall anywhere in the range from 10 g to 111 g, and that is believed to be the range at which concussions could potentially occur based on the football-related incidents \cite{bakhos2010emergency}. Furthermore, concussion could occur at a low peak linear acceleration of 31.8 g as reported by another study \cite{mihalik2010collision}. Another study analyzed 62,974 head acceleration data points and identified subconcussive impacts with an average head acceleration of 26 $\pm$ 20 g with a median of 19 g while the average head acceleration for concussive impacts was 105 $\pm$ 27 g with a median of 103 g \cite{rowson2011development}.

Laboratory settings using dummies are commonly used to evaluate the potential harm to a human by simulating a certain scenario, such as that used to evaluate harm levels due to impact in industrial robots \cite{haddadin2008role}. Similar setups have been adopted in other fields to quantify harm due to impacts, such as that in sports to evaluate potential harm and the efficacy of protective gears \cite{mcintosh2003evaluation}\cite{yang2014baseball}\cite{o2013safety}. Similarly, a laboratory setup generating the peak linear head acceleration severity index will be used in this study as an indicator to understand the effects on the head linear acceleration due to altering some of the parameters of the impacting objects.

\subsection {Data Analysis}
Two main analyses were conducted on the generated results. The first analysis was a one-way analysis of variance (ANOVA). This test was conducted on the average responses of the peak head acceleration at each level of each factor to understand the contribution of each factor on the resultant response. The level for statistical significance was set to \textit{p} $<$ 0.05.

The second analysis was based on the signal-to-noise (S/N) ratio and it was used for optimization. The goal of this analysis is to minimize the response variable, hence, the smaller-the-better case was considered and the corresponding S/N ratio was used; it is defined as follows \cite{cavazzuti2012optimization}:

\begin{equation}
S/N = -10  log_{10}E[y_{i}^{2}]
\end{equation}
where $\textit{E}$ is the expected value and \textit{$y_{i}^{2}$} is the response. All analyses were performed using Minitab (v17.0, Minitab Inc., USA).

\section{Results}
\subsection{Orthogonal array}
A total of 27 responses of the peak linear head acceleration were recorded and the corresponding average value, standard deviation, and S/N ratio for each combination were calculated to complete the Taguchi $L_{9}$ orthogonal array (Table \ref{complete_OA}). The obtained linear acceleration values were in the range from 2.42 to 10.75 g due to different levels of control and noise factors. The lowest linear acceleration value obtained corresponds to a thickness of 3 mm, Ecoflex, and Level 1 impact velocity (i.e. A2-B1-X1) while the highest linear acceleration value corresponds to a thickness of 1 mm, Ecoflex, and Level 3 impact velocity (i.e. A1-B1-X3). The average response and the S/N ratio due to varying the level of each factor were tabulated (Table \ref{means_only} and Table \ref{SNR_only}). The lowest average linear head acceleration was 3.18 g with S/N ratio of -10.10 occurred at Level 1 impact velocity while the highest average linear head acceleration was 10.10 g with S/N ratio of -20.09 due to Level 3 impact velocity.

\begin{table*}[!]
	\centering
	\caption{The complete Taguchi orthogonal array along with columns showing the average response, standard deviation (SD), and the signal to noise ratio (S/N) for each row.}
	\label{complete_OA}
	\begin{tabular}{ccccccccc}
		\toprule
		& \multicolumn{2}{p{1.5cm}}{\centering Inner control \\ factors array} & \multicolumn{3}{p{2.5cm}}{\centering Outer noise factor \\ array} &\multicolumn{1}{p{1.5cm}}{\centering Average\\ response}& \multicolumn{1}{p{1.5cm}}{\centering Standard\\ deviation} & \multicolumn{1}{p{1.5cm}}{\centering Signal-to-noise \\ ratio }\\\midrule
		RUN& A & B & X1 & X2 & X3 & Mean & SD & S/N \\
		1 & 1 & 1 & 2.90 & 6.56 & 10.75 & 6.74 & 3.92 & -17.45 \\
		2 & 1 & 2 & 3.98 & 6.04 & 10.23 & 6.75 & 3.18 & -17.19 \\
		3 & 1 & 3 & 3.41 & 6.35 & 10.26 & 6.67 & 3.43 & -17.19 \\
		4 & 2 & 1 & 2.42 & 5.65 & 9.99 & 6.02 & 3.80 & -16.61 \\
		5 & 2 & 2 & 3.24 & 6.01 & 10.29 & 6.51 & 3.55 & -17.06 \\
		6 & 2 & 3 & 3.13 & 6.06 & 10.18 & 6.46 & 3.54 & -16.99 \\
		7 & 3 & 1 & 3.03 & 6.15 & 10.03 & 6.40 & 3.50 & -16.92 \\
		8 & 3 & 2 & 3.38 & 6.14 & 10.53 & 6.68 & 3.60 & -17.27 \\
		9 & 3 & 3 & 3.08 & 6.50 & 8.59 & 6.06 & 2.78 & -16.21    \\  
		\bottomrule   
	\end{tabular}
\end{table*}

\begin{table}[!]
	\centering
	\caption{The average peak head acceleration for each factor at every level. }
	\label{means_only}
	\begin{tabular}{>{\centering\arraybackslash}p{1.65cm}ccc}
		\toprule
		& A & B & X \\
		Level & Mean (SD)&  Mean (SD) & Mean (SD) \\\midrule
		1 & 6.72 (0.04)  & 6.39 (0.36)  & 3.18 (0.42)  \\
		2 & 6.33 (0.27)&  6.65 (0.12)&  6.16 (0.28) \\
		3 & 6.38 (0.31)&  6.40 (0.31)&  10.10 (0.61) \\\bottomrule
	\end{tabular}
\end{table}

\begin{table}[!]
	\centering
	\caption{The S/N values for each factor at every level. }
	\label{SNR_only}
	\begin{tabular}{>{\centering\arraybackslash}p{1.7cm}>{\centering}p{1.75cm}>{\centering}p{2.5cm}cp{2cm}}
		\toprule
		& A & B & X \\
		Level &  S/N &  S/N &  S/N \\\midrule
		1 &  -17.28 &  -16.99 &  -10.10 \\
		2 &  -16.89 &  -17.17 &  -15.80 \\
		3 &  -16.8 &  -16.80 & -20.09\\\bottomrule
	\end{tabular}
\end{table}

\subsection{ANOVA Tests}
One-way ANOVA tests for each factor were conducted to understand if there is a significant difference due to varying the conditions on the resultant average peak linear head acceleration. In case the ANOVA test reported a significant difference, a post-hoc Tukey test was conducted.
\subsubsection{Effect of the material's thickness}
A one-way ANOVA was conducted to compare the effect of varying the three conditions of the thickness on the response (Table \ref{ANOVA_MASS}). The test revealed that there was no significant difference due to varying the thickness on the resultant average linear head acceleration for the three conditions (F(2,24) = 0.04, \textit{p} = 0.96) at the \textit{p}$<$ 0.05.
\begin{table}[!]
	\centering
	\caption{ANOVA analysis for the effect of material's thickness on the linear head acceleration}
	\label{ANOVA_MASS}
	\begin{tabular}{p{0.775cm}p{0.4cm}p{1.5cm}p{1cm}cc}
		\toprule
		Source & df & Sum of squares & Mean square & F-Value & P-Value \\\midrule
		Factor & 2  & 0.81         & 0.41      & 0.04    & 0.96   \\
		Error  & 24 & 220.854         & 9.20       &         &         \\
		Total  & 26 & 221.67        &             &         &        \\\bottomrule
	\end{tabular}
\end{table}

\subsubsection{Effect of the material's storage modulus}
A one-way ANOVA was conducted to compare the effect of the three different conditions of the storage modulus on the head peak acceleration (Table \ref{ANOVA_SHAPE}). The test revealed that there was no significant difference due to changing the storage modulus on the resultant average linear head acceleration for the three conditions (F(2,24) = 0.02, \textit{p} = 0.98) at \textit{p}$<$ 0.05.

\begin{table}[!]
	\centering
	\caption{ANOVA analysis for the effect of material's storage modulus on the linear head acceleration}
	\label{ANOVA_SHAPE}
	\begin{tabular}{p{0.775cm}p{0.4cm}p{1.5cm}p{1cm}cc}
		\toprule
		Source & df & Sum of squares & Mean square & F-Value & P-Value \\\midrule
		Factor & 2  & 0.39          & 0.20       & 0.02    & 0.98   \\
		Error  & 24 & 221.27        & 9.22       &         &         \\
		Total  & 26 & 221.67        &             &         &      \\\bottomrule  
	\end{tabular}
\end{table}

\begin{table}[!]
	\centering
	\caption{ANOVA analysis for the effect of impact velocity on the linear head acceleration}
	\label{ANOVA_SPEED}
	\begin{tabular}{p{0.775cm}p{0.4cm}p{1.5cm}p{1cm}cc}
		\toprule
		Source & df & Sum of squares & Mean square & F-Value & P-Value \\\midrule
		Factor & 2 & 216.63 & 108.31 & 515.63 & 0.00 \\
		Error & 24 & 5.04 & 0.21 & & \\
		Total & 26 & 221.67 & & &\\\bottomrule
	\end{tabular}
\end{table}

\subsubsection{Effect of the object's impacting velocity}
A one-way ANOVA was conducted to compare the effect of the three different levels of the impact velocity on the head peak acceleration (Table \ref{ANOVA_SPEED}). The test revealed that there was a significant difference due to varying the impact velocity on the resultant average linear head acceleration for the three conditions (F(2,24) = 515.63, \textit{p} = 0.00) at the \textit{p}$<$ 0.05. A post-hoc Tukey test showed that velocity Level 1 (M = 3.18, SD = 0.42), velocity Level 2 (M = 6.16, SD = 0.28), and velocity Level 3 (M = 10.10, SD = 0.61) were different significantly at \textit{p}$<$ 0.05.

\section{Discussion}
\subsection{Analysis}
The alteration of the control factors and noise factor levels have an effect on the resultant head accelerations (Table \ref{complete_OA}). No definite trend can be observed between the thickness  (i.e. Factor A) and the resultant response by visually investigating the orthogonal array. For example, looking at the response values in the noise factor columns X1 - X3 from row 1 to 9, the registered head acceleration value appear to remain consistent. Similar observation can be made for Factor B. One the other hand, the noise factor levels seem to affect the response significantly. For example, examining columns X1, X2, and X3 reveals that the response increases proportionally with the applied impact velocity (i.e. noise factor) as supported by the relatively large standard deviations across each row.

The peak linear head acceleration has been reported to be influenced by the impact velocity of an impactor \cite{hutchinson2014peak}\cite{haddadin2008role}. The increasing average values of the response at each level of the impact velocity supports these findings (Table \ref{means_only}). Furthermore, ANOVA test results and post-hoc Tukey findings showed that the impact velocity has a significant effect on the resultant peak head acceleration. On the other hand, no similar conclusion can be made for the control factors. For example, the reported ANOVA results for the effect of the material's thickness revealed that there was no statistical significance. This could be attributed to the relatively small selected thickness range and to the small difference between each level.

\subsection{Optimization}
\begin{figure*}[!]
	\centering
	\includegraphics[width=1.0\textwidth]{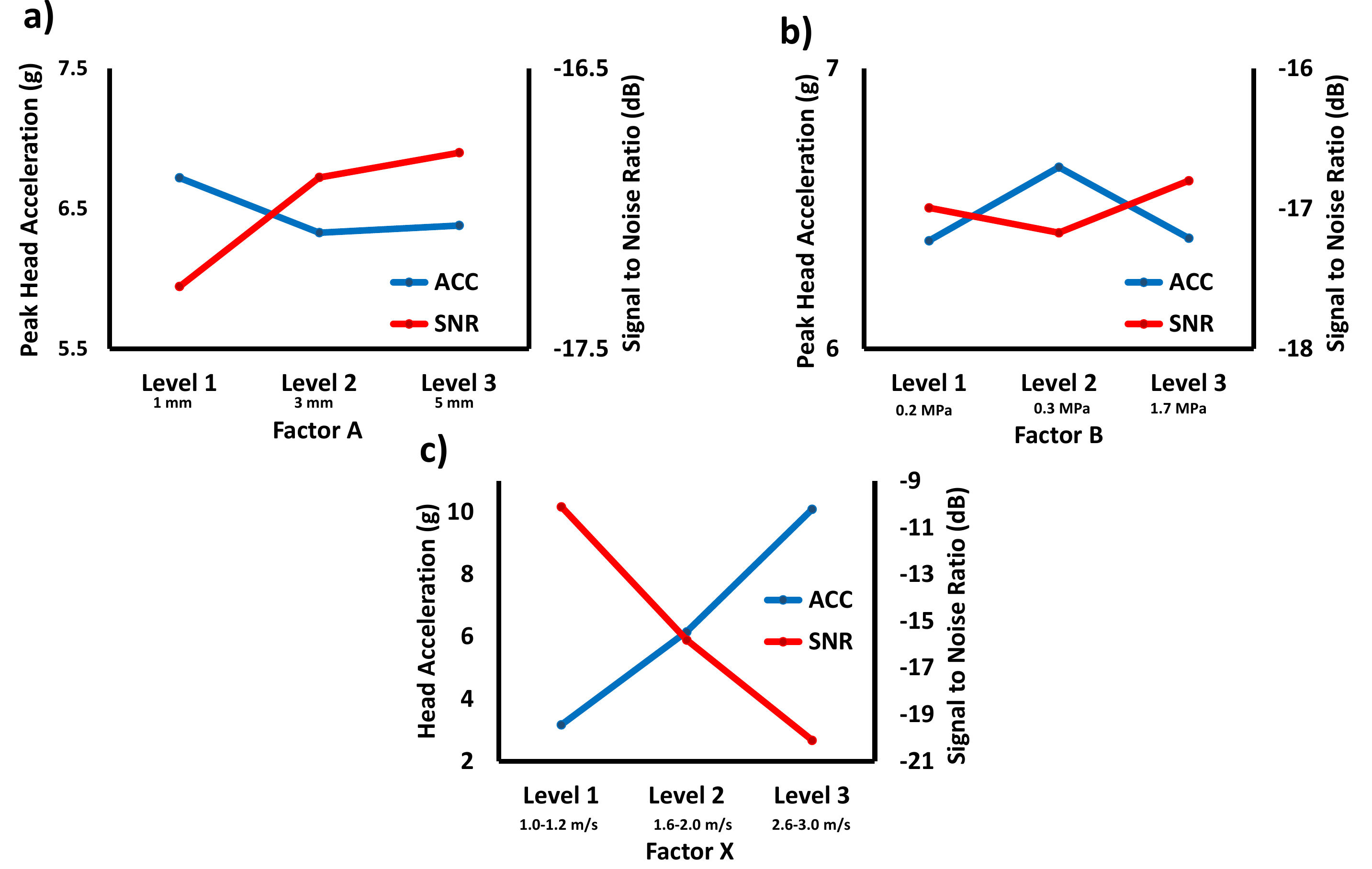}
	\caption{The resultant peak head acceleration and S/N ratio for each factor considered in this study. a) For Factor A, the material thickness. b) For Factor B, the storage modulus. c) For Factor X, the impact velocity.}
	\label{SNR_fig}
\end{figure*}

The second goal of this study is to find the optimal values that reduces the response. This is achieved by investigating the mean value of the resultant head acceleration and the corresponding mean S/N ratio for each factor (Table \ref{means_only} and Table \ref{SNR_only}). The plots were generated for the mean response at each factor level and the corresponding mean S/N ratios for a better visual comparison (Fig. \ref{SNR_fig}). The criterion for selecting the optimal conditions is based on finding the levels that produce the lowest response and highest S/N ratio.

The best conditions for the thickness and storage modulus were identified based on the lowest generated head linear acceleration and highest S/N ratio (Fig. \ref{SNR_fig}). For the control factor of material's thickness (i.e. Factor A), 3 mm (i.e. Level 2) and 5 mm (i.e. Level 3) achieved closely the best results. As for control factor of material's storage modulus (i.e. Factor B), ecoflex (i.e. Level 1) and clay (i.e. Level 3) scored closely the best results. Even though ANOVA tests on both of the control factors reported no significance in affecting the peak head acceleration, the identified levels for each factor provided the lowest responses and highest S/N ratios as compared to other conditions. Hence, these conditions were selected as the optimal values. As for the noise factor, 1 - 1.2 m/s impact velocities (i.e Level 1) scored the best results. The optimized conditions for the control factors investigated were 3 mm or 5 mm for the material's thickness factor and ecoflex or clay for the material's storage modulus (i.e. A2-B1, A2-B3, A3-B1, and A3-B3). Interestingly, the selected optimal control factors' levels produced relatively lower average peak head accelerations compared to other conditions at even higher noise levels (Table \ref{complete_OA}).

\begin{table}[!]
	\centering
	\caption{The results of the confirmation runs that were conducted at the optimal identified conditions}
	\label{Conf_runs}
	\begin{tabular}{p{2cm}p{2cm}p{2cm}p{2cm}}
		\toprule
		Combination            & X1   & X2   & X3   \\\midrule
		\multirow{3}{*}{A2-B1} & 2.17 & 5.11 & 10.18 \\
		& 3.17 & 6.74 & 9.45 \\
		& 2.68 & 5.60 & 9.69 \\
		Mean (SD)    & 2.67 (0.50) & 5.81 (0.83) & 9.78 (0.37)\\\midrule
		\multirow{3}{*}{A2-B3} & 2.12 & 5.04 & 9.18 \\
		& 3.43 & 6.59 & 10.60 \\
		& 3.37 & 5.06 & 10.25 \\
		Mean (SD)    & 2.97 (0.74) & 5.56 (0.89) & 10.01 (0.74)\\\midrule
		\multirow{3}{*}{A3-B1} & 3.06 & 6.27 & 10.25 \\
		& 2.27 & 6.13 & 9.71 \\
		& 2.87 & 6.27 & 9.82 \\
		Mean (SD)    & 2.74 (0.41) & 6.22 (0.08) & 9.93 (0.28)\\\midrule
		\multirow{3}{*}{A3-B3} & 3.34 & 7.08 & 8.54 \\
		& 2.99 & 6.38 & 8.76 \\
		& 3.32 & 6.4 & 8.60 \\
		Mean (SD) & 3.22 (0.20) & 6.62 (0.40) & 8.63 (0.11) \\\bottomrule
	\end{tabular}
\end{table}

\subsection{Confirmation Tests}

After finding the optimal levels for each control factor, the last stage of Taguchi design is to perform the confirmation runs. The goal of this study is to minimize the peak head acceleration due to an object being thrown at the head by investigating the effect of two control factors, namely the material's thickness and its storage modulus. Hence, the optimal levels obtained in the previous section should produce relatively smaller head accelerations as compared to other conditions. Confirmation runs are needed to confirm these findings. To ensure that the optimal levels are robust and applicable to different noise scenarios, two confirmation runs will be conducted at every noise level.  

A total of 24 confirmation runs were conducted at the optimal control factors' levels. For each control and noise factors combination, 2 runs were conducted and the corresponding mean values for each were calculated (Table \ref{Conf_runs}). Comparing the results of the confirmation runs to that obtained from the main experiments, the average values were very close to respective ones obtained in the complete Taguchi orthogonal array (Table \ref{complete_OA}). Hence, the confirmation runs confirmed that the selected optimal levels produced the lowest peak head accelerations.

\subsection{Limitations of the Study}
This study considered only the application of three soft materials while there are many other candidates that could be considered, for example, PDMS \cite{Husain2016}. The effects of the added mass of the soft materials were ignored (e.g Less than 0.05 kg). However, this added mass might influence the results significantly, especially when larger area is covered (e.g. covering the whole object with a soft material) or larger thickness is considered (i.e. greater than 5 mm). For consistency, the shape of the object was limited to one shape while the velocity of impacts was limited to low range. However, different shapes of robotic toys exist and higher impact velocities might occur in realistic scenarios. Other severity indices could have been considered to measure different potential harm. For example, measuring the soft tissue injuries and quantify the potential of soft materials in mitigating it.

\section{Conclusion}

In this study, the influence of an added soft material to an object on the linear acceleration of the head upon impact has been investigated. The Taguchi $L_{9}(3^{2})$ orthogonal array design has been used to plan the 27 main experiments that were conducted. The control factors were the thickness and the storage modulus of three different soft materials. The noise factor was the impact velocity. The significance of each factor has been identified based on ANOVA tests while the optimal levels for the control factors were identified based on the analysis of S/N ratio. ANOVA tests showed that the control factors were not statistically significant in influencing the linear acceleration of the head. On the other hand, ANOVA test of the noise factor revealed that it was statistically significant. Material thickness of 3 mm and 5 mm achieved the best results. This implies that the application of a higher thickness of soft material will attenuate the head's acceleration better. Ecoflex and clay have achieved better response as compared to dragon skin. Confirmation runs at the optimal identified conditions achieved better responses as compared to other conditions.        

The control factors, especially for the thickness of a soft material, may provide reduction in the overall head's acceleration. Hence, the manufacturers and designers of small robotic toys for special needs children should consider adding a layer of safe and soft material to their products to improve safety and to reduce any potential harm (e.g. subconcussions and superficial injuries). Additionally, they need to investigate different soft materials to find suitable materials that provide robustness and ease of application to their products while improving the safety aspects. Considering soft materials will also open for the possibility of embedding different sensors \cite{sadasivuni2017strain}\cite{cabibihan2019biometric}\cite{Alhaddad2017toward}. The investigation conducted in this study used Taguchi method to design the experiments in which it provided a convenient, cost effective, and efficient way in the assessment and optimization of product designs concerning safety aspects. The emerging field of companion social robots can benefit from Taguchi methods in its approach to optimize a robot$'$s design.

\section*{Acknowledgements}
The work is supported by a research grant from Qatar University under the grant No. QUST-1-CENG-2019-10. The statements made herein are solely the responsibility of the authors.

\bibliographystyle{unsrt}  
%\bibliography{refer_mat}


\end{document}